\documentclass[a4paper, 10pt, conference]{ieeeconf}      
\usepackage{FG2025}
\usepackage{graphicx}
\FGfinalcopy 

\IEEEoverridecommandlockouts                              
\def\FGPaperID{****} 

\title{\LARGE \bf
Unconstrained Body Recognition at Altitude and Range: Comparing Four Approaches
}

\author{\parbox{16cm}{\centering
    {\large Blake A Myers, Matthew Q Hill, Veda Nandan Gandi, Thomas M Metz, Alice J O'Toole}\\
    {\normalsize
    School of Behavioral and Brain Sciences, The University of Texas at Dallas, Richardson, Texas\\}}
}

\begin{document}

\ifFGfinal
\thispagestyle{empty}
\pagestyle{empty}
\else
\author{Anonymous FG2025 submission\\ Paper ID \FGPaperID \\}
\pagestyle{plain}
\fi
\maketitle

\begin{abstract}

This study presents an investigation of four distinct approaches to long-term person identification using body shape. Unlike short-term re-identification systems that rely on temporary features (e.g., clothing), we focus on learning persistent body shape characteristics that remain stable over time.
We introduce a body identification model based on a Vision Transformer (ViT) (Body Identification from Diverse Datasets, BIDDS) and on a Swin-ViT model (Swin-BIDDS). We also expand on previous approaches \cite{myers2023recognizing} based on the Linguistic and Non-linguistic Core ResNet Identity Models (LCRIM and NLCRIM), but with improved training.
All models are trained on a large and diverse dataset of over 1.9 million images of approximately 5k identities across 9 databases. Performance was evaluated on standard re-identification
benchmark datasets (MARS \cite{zheng2016mars}, MSMT17 \cite{wei2018person}, Outdoor Gait \cite{song2017learning}, DeepChange \cite{xu2023deepchange}) and on an unconstrained dataset \cite{cornett2023expanding} that includes images
at a distance (from close-range to 1000m),  at altitude 
(from an unmanned aerial vehicle, UAV), and with clothing change. 
A comparative analysis across these models
provides insights into how different backbone architectures and input image sizes
impact long-term body identification
performance across real-world conditions.
\end{abstract}

\section{INTRODUCTION}

Face recognition algorithms are highly accurate at establishing the unique identity of individuals (e.g., \cite{deng2019arcface, kim2022adaface, wang2018cosface}). In natural viewing conditions, however, facial identity information is commonly degraded or obscured (e.g., viewing from a far distance or at an extreme angle).
When the face is unusable or inaccessible, information about the shape of the body can constrain identity decisions. Body shape can contribute to person identification by supporting/vetoing uncertain face identifications and/or by establishing a plausible identity match to a gallery image. As such, it can serve as a valuable biometric, even if it provides information that does not uniquely identify an individual.

Early attempts to use the body for identification focused on re-identification
in a closed-world setting, which aims to
track a person in a constrained environment like an airport or train station (for a review, see \cite{9336268}). In a closed-world environment, algorithms 
can rely on cues such as clothing that make the problem
easily tractable with deep learning. 
In an open-world, 
clothing change makes the problem  more difficult
\cite{chen2021learning, gu2022clothes, hong2021fine, huang2023whole, liu2023learning, liu2024distilling, myers2023recognizing, yang2019person, ye2021deep, myers2023recognizing, liu2024distilling, huang2023whole}. As algorithms
capable of overcoming clothing change have matured (for a review, see \cite{9336268}),
long-term body identification models have
aimed more broadly at identification in highly challenging scenarios  \cite{huang2023whole, huang2023selfsupervised, huang2024villsvideoimagelearninglearn, myers2023recognizing}. This emphasis coincides with the development of the BRIAR dataset, which contains whole person images, captured at long-range (e.g., 300+ meters), through atmospheric turbulence, and/or from elevated sensor platforms \cite{cornett2023expanding}. Images and videos in this dataset are taken across multiple views (yaw, pitch) and with clothing changes that sideline short-term cues that are useful in re-identification scenarios \cite{yi2014learning}.   

Long-term body identification with  unconstrained data presents a number of unique challenges. Body shape can deform via the movements of the limbs (e.g., arms up or down, leg extended) and/or by changes in posture (e.g., bending, reaching). Texture and albedo information, which are critically important for face identification, have only limited value for clothed bodies. Perhaps most challenging is the complicating factor of clothing change that alters the overall shape of the body (e.g., pants/skirt, running gear/winter coat). Despite the difficulty of the task, there is ample evidence that humans use body information for person identification when the face is unavailable or insufficiently resolved for identification \cite{hahn2016dissecting, rice2013role, yovel2016recognizing}.  

The goal of the present study was to compare four machine learning approaches to real-world (longer-term) body identification. We evaluated two Vision Transformer models (ViT) \cite{dosovitskiy2020image, liu2021swin} and two ResNet neural networks  \cite{he2015deep}. All four models were trained to identify bodies from input images, using a very large dataset 
compiled from 9 feeder datasets.  
The use of a common dataset for training allowed us to compare the models on an equal footing. 

In the first part of the work, we compared the models  on four common re-ID datasets.
In the second part of the work, we tested  the
models on the BRIAR test set (BTS) \cite{cornett2023expanding}.  We also test performance
on subsets of data that measure body identification
with face included and restricted, at long range,
and from overhead. In the third part of the work, 
we dissect the advantage of the best model to determine whether architecture or image size accounts for its superiority.

The contributions of the paper are:
\begin{itemize}
     \item We show  that ViT models are superior to equivalently-trained ResNet models for body identification.  
     \item We show that a Swin-ViT model is superior to the other tested models across metrics. This was true for  the benchmark datasets and the unconstrained dataset.
     \item Linguistic pre-training in a ResNet model showed only a small performance advantage over an equivalent non-linguistically trained model.
     \item We show that  both architecture and image size contributed to the superior performance of the Swin-ViT, but that image size was the critical factor in its high performance.
          

\end{itemize}

\subsection{Related Work}

Long-term body identification models  can be categorized according to the approach they take 
to representing body shape information.

\subsubsection{2D Body Shape from Images}
The most direct approach is to learn
a mapping from variable images of bodies
(view, clothing, illumination, distance) to identity
\cite{han2023clothing,myers2023recognizing}.
The greatest challenge in this approach has been
the limited availability of training data with sufficient
variability (especially clothing sets) to learn the task
of long-term body identification.
Using a ResNet-50 model pretrained on ImageNet \cite{russakovsky2015imagenet}, the Clothing-Change Feature Augmentation (CCFA) approach  \cite{han2023clothing} augments
model training to form meaningful clothing variations in the feature
space. The augmented features maximize
the change of clothing and minimize the change of identity
by adversarial learning.
The effectiveness of CCFA was demonstrated with 
two standard CC-ReID datasets (PRCC-ReID \cite{yang2019person} and  LTCC-ReID \cite{qian2020long}).

   \begin{figure}[thpb]
      \centering
      \includegraphics[scale=0.85]{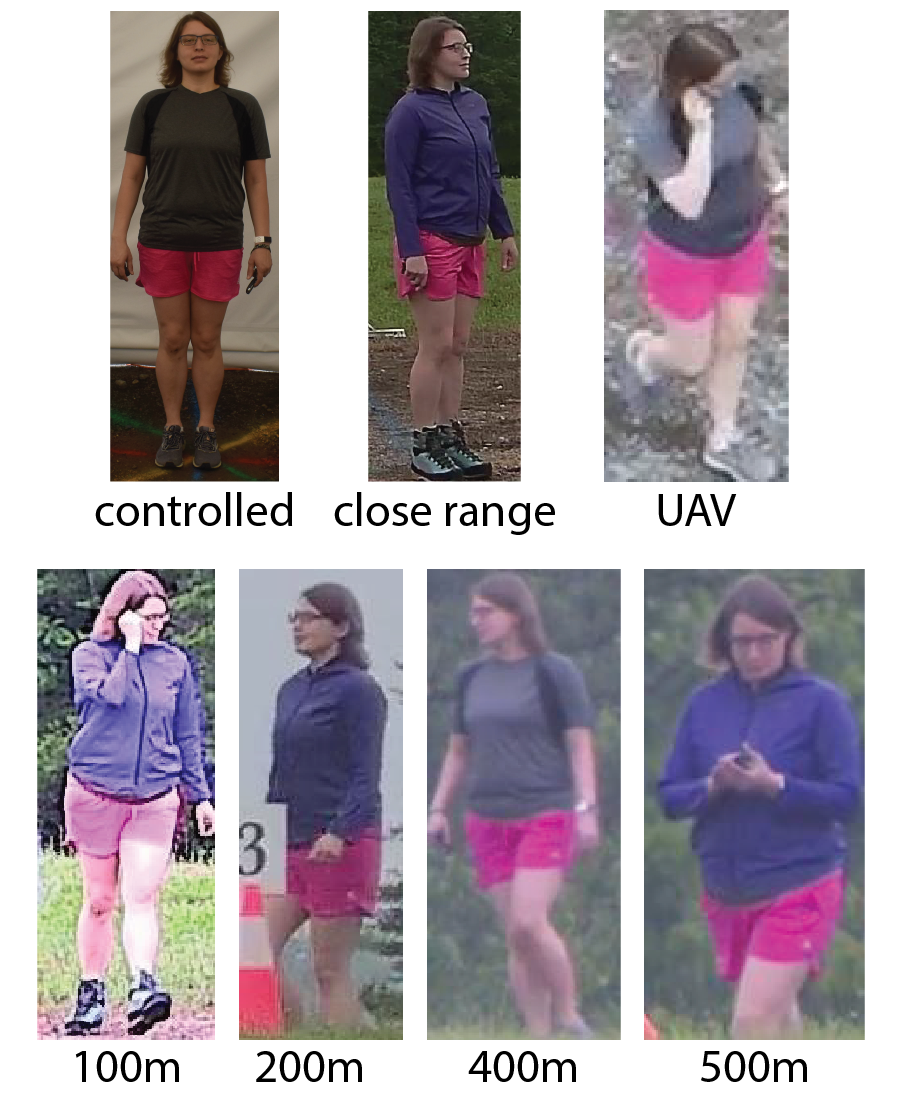}
      \caption{Example body images from the BTS dataset \cite{cornett2023expanding}. Subject consented to publication. }
      \label{BRIAR_images}
      \vskip -0.25cm
   \end{figure}

 The Non-linguistic Core ResNet Identity Model (NLCRIM) \cite{myers2023recognizing} was built on a ResNet-101 backbone pretrained with ImageNet \cite{russakovsky2015imagenet}.  NLCRIM was 
trained to map body images to identities using the BRIAR Research Set (BRS) 
 \cite{cornett2023expanding}. It was  
evaluated with the BRIAR Test Set (BTS), which contained
identities viewed at multiple distances (up to 1000 meters)
that varied widely in yaw and pitch. Extreme pitch conditions
were captured from unmanned aerial vehicles (UAVs). All
test items included a change of clothing.
(See Figure \ref{BRIAR_images} for image examples). 
NLCRIM performed well
across all probe distance/pitch conditions. 
An improved version
of this model, with enhanced training 
and  substantially more training data, is tested in the present
study.

A similar direct approach to learning a mapping between 
whole body images and identity was taken in \cite{huang2024whole}.  A ResNet-50 model was trained from scratch
with BRS data. This body encoder was embedded
in an end-to-end system that included a trained detector
model. The combined model performed well on the unconstrained BTS data.

The causality-based autointervention model (AIM1) was
proposed to mitigate clothing bias for robust clothes-changing person ReID
(CC-ReID) \cite{yang2023good}. Specifically, 
 the effect of clothing
on model inference was analyzed. A dual-branch structure of clothing and ID was
utilized to simulate the causal intervention process and 
was penalized by a causality loss. Progressively, clothing bias
was  automatically eliminated with model training, as AIM
 learned more discriminative identity clues that are
independent of clothing. The  superiority of the
 AIM approach over other approaches was demonstrated
 with two standard CC-ReID datasets (PRCC-ReID \cite{yang2019person} and  LTCC-ReID \cite{qian2020long}).

\subsubsection{3D Body Shape Features}
To overcome reliance on short-term cues in body images,
several models have attempted to reconstruct 3D body shapes for
identification (\cite{chen2021learning, liu2023learning}).
In the 3D Shape Learning (3DSL) approach, a texture-insensitive 3D shape embedding is extracted
from a 2D image by adding 3D body reconstruction as an
auxiliary task and regularization \cite{chen2021learning}. 
The use of the 3D reconstruction regularization
forces a  decoupling of the 3D body shape from the visual texture, 
enabling the model to acquire discriminative 3D
shape ReID features.  An adversarial self-supervised 
projection (ASSP) model is used to provide a 3D shape ground truth. The effectiveness of the approach was
demonstrated with common person ReID
datasets (e.g., Market1501 \cite{zheng2015scalable}) and clothes-changing datasets (e.g., PRCC-ReID \cite{yang2019person} and  LTCC-ReID \cite{qian2020long}).

In other work, the 3DInvarReID model \cite{liu2023learning} begins by
disentangling identity from
non-identity components (pose, clothing shape, and texture)
of 3D clothed humans. Next, 
 accurate 3D
clothed body shapes are reconstructed,
and discriminative  features
of naked body shapes for person ReID are learned. The model was found to be effective for disentangling identity and non-identity features in 3D clothed
body shapes,  using a dataset (CCDA \cite{liu2023learning}) that contains a wide variety of human activities and clothing changes.

\subsubsection{Linguistic Models} 
Body models based on linguistic descriptors  (e.g., “curvy,” “long-legged”)  encode  shape via the complex myriad of features captured by single and small groups of words \cite{myers2023recognizing}. 
Work in psychology \cite{hill2016creating} and computer graphics \cite{streuber2016body} has  demonstrated that a linear mapping can be learned from human-generated body descriptions (27 words) to the coefficients of a PCA trained with 3D body scans \cite{loper2015smpl}. Motivated by this finding, the Linguistic Core ResNet Identity Model (LCRIM)
 was developed using an ImageNet pretrained ResNet augmented
with linguistic annotation pretraining. This linguistic core was then trained to map images to identity \cite{myers2023recognizing}. Although the LCRIM model performed at a level similar to NLCRIM, the fusion of the two models performed substantially better than either model alone. This suggests that the two models encode complementary 
information about body shape.

In related work, linguistic body descriptions were
leveraged for ReID in CLIP3DReID \cite{liu2024distilling}. This was done by integrating human descriptions with visual perception using a pretrained CLIP model. CLIP was used to automatically label body shapes with linguistic descriptors. A student model's local visual features were then aligned with shape-aware tokens derived from CLIP's linguistic output. The CLIP image encoder and the 3D SMPL \cite{loper2015smpl} identity spaces were used in combination to align the global visual features. The effectiveness
of CLIP3DReID was demonstrated using  PRCC-ReID \cite{yang2019person} and  LTCC-ReID \cite{qian2020long}.


\section{Methods}\label{sec:Methods}
\subsection{Model Architectures}

We examined four distinct approaches to body shape recognition: two vision transformer models (BIDDS and Swin-BIDDS) and two ResNet-based models (LCRIM and NLCRIM) \cite{myers2023recognizing}. Each architecture employs a unique strategy for capturing body shape features across varying conditions.


\subsubsection{Vision Transformer Models}
The BIDDS model is built on a Vision Transformer architecture. We used  a ViT-B/16  pre-trained on ImageNet-1k. The core model processes $224\times224$ sized images with patch size 16. We modified the original ViT architecture by replacing the classification head with a custom fully connected layer that maps to a 2048-dimensional embedding. This embedding space is designed to capture essential body shape features crucial for person identification. Following core training, we fine-tune the model on the BRS1--5 datasets (cf. \cite{cornett2023expanding}), increasing image size to $384\times384$ to capture more detailed features
of the fine-tuning BRIAR data, while maintaining the same architectural structure.

Swin-BIDDS is based on the hierarchical vision transformer, which 
uses shifted windows (Swin Transformer, 
\cite{liu2021swin}). This type of transformer
 was developed to better adapt transformers from the language domain to
 the vision domain, by accommodating large variations in the scale of visual entities. The shifted windowing scheme of the Swin Transformer is more efficient than a standard ViT,
 because it limits self-attention computation to non-overlapping local windows,  while supporting cross-window connections. This hierarchical structure 
progressively merges patches and is well-suited to
 modeling at various scales. 
\subsubsection{ResNet-Based Models}

These models leverage the ResNet architecture with different core training strategies. Both are
pre-trained with ImageNet-1k \cite{russakovsky2015imagenet}. Additionally, 
 LCRIM incorporates semantic body descriptors into its training process (See Section \ref{LCRIM_pretraining} for details). 
Its architecture consists of a ResNet-50 base
augmented with an encoder-decoder structure that maps to a linguistic feature space before the final identification layers. The encoder pathway compresses the representation (2048 → 512 → 64 → 16), while the decoder pathway (16 → 24 → 30) reconstructs linguistic body attributes.  
NLCRIM is identical to LCRIM, but without  linguistic training.  

By comparison
to the published version of NLCRIM and LCRIM \cite{myers2023recognizing}, there were three changes: 1.)
a new training regime with hard triplet mining was added \cite{hermans2017defense}; 2.) there was a substantial increase
in the quantity of training data; and 3.) the ResNet-101 was replaced by a ResNet-50 (see below for details).

\subsection{Training Methods}

All models employed hard triplet loss with negative mining \cite{hermans2017defense}. This operates on image triplets: an anchor image, a positive sample (same identity), and a negative sample (different identity). The loss calculation measures the Euclidean distances between the anchor and positive samples and between the anchor and negative samples. We selected the most challenging negative samples (i.e., those closest to the anchor in the embedding space) within each batch. This hard negative mining encourages the model to learn features that effectively differentiate between similar body shapes. We also ensured that each batch  included pairs or small sets of images from the same person. 
All four models use the Adam optimizer and incorporate dynamic sampling, whereby triplet selection is adapted based on the current state of the embeddings. This ensures that the models continuously encounter challenging examples throughout training. The training process employs a low learning rate ($10^{-5}$) and weight decay ($10^{-6}$) to prevent over-fitting while maintaining  stability.
We applied standard augmentations during training, including
random horizontal flip, color jitter, random grayscale, and gaussian blur.



\begin{table}[ht]
\center
\caption{Training and Testing Datasets}
\vspace{10pt}
\label{combined_datasets}
\begin{tabular}{|p{2.6cm}||p{1.25cm}||p{1.25cm}||p{1.5cm}||}
\hline
\textbf{Dataset} & \textbf{Images} & \textbf{IDs} & \textbf{Clothes Change}  \\
\hline
UAV-Human \cite{9578530} & 41,290 & 119  & no  \\\hline
MSMT17 \cite{wei2018person} & 29,204 & 930  & no  \\\hline
Market1501 \cite{zheng2015scalable} & 17,874  & 1,170  & no  \\\hline
MARS \cite{zheng2016mars} & 509,914 & 625  & no   \\\hline
STR-BRC 1 & 156,688 & 224  & yes  \\\hline
P-DESTRE \cite{kumar2020p} & 214,950  &  124  & no  \\\hline
PRCC \cite{yang2019person} & 17,896 & 150  & yes  \\\hline
DeepChange \cite{xu2023deepchange} & 28,1731 &  451  & yes \\\hline
BRS 1--5 \cite{cornett2023expanding} & 697,348  & 995  & yes   \\\hline
\textbf{Total Training} &  1,966,895 & 4,788  &\\\hline
\hline
MSMT17 \cite{wei2018person} & 82,510 & 2,697  & no  \\\hline
MARS \cite{zheng2016mars} & 509,966 & 634  & no   \\\hline
Outdoor Gait \cite{song2017learning}  & 402,009  &  138  & no  \\\hline
DeepChange \cite{xu2023deepchange} & 103,324 &  671  & yes  \\\hline
\textbf{Total Testing} &  1,097,809 & 4,140 &  \\\hline
\end{tabular}
\label{training_datasets}
\end{table}


\subsection{Training Data}
An important feature of our approach
is the use of a large and diverse collection of training
datasets  (see
Table \ref{training_datasets}).  The datasets include
over 1.9 million images across 4,788
identities. To benchmark the models for the experiments
(see Section \ref{expt_section}), a subset of test data were withheld from three of the training sets (MSMT17 \cite{wei2018person}, MARS \cite{zheng2016mars}, and DeepChange \cite{xu2023deepchange}), and a fourth set was designated solely for testing purposes (Outdoor Gait \cite{song2017learning}), with none of its data included in the training phase.
This diverse collection spans multiple scenarios, from ground-level views to aerial perspectives. The training images were primarily derived from video files, with bodies cropped and processed to maintain their aspect ratios while being placed on a $224\times224$ 
($384\times384$ for Swin-BIDDS) black background. Some of the
datasets include clothing change and some do not
(see Table \ref{training_datasets}).


\begin{table*}
\caption{Test set performance.}
\begin{center}
\begin{tabular}{|l|l|c|c|c|c|c|}
\hline
Dataset & Model & AUC & TAR@FAR $10^{-3}$ & TAR@FAR $10^{-4}$ & Rank 1 & Rank 20 \\
\hline\hline
MARS & NLCRIM & 0.9942 & 0.8059 & 0.4621 & 0.8868 & 0.9774 \\
 & LCRIM & 0.9962 & 0.9235 & 0.7468 & 0.9147 & 0.9785 \\
  & BIDDS & 0.9958 & 0.9509 & 0.8562 & 0.9476 & 0.9824 \\
 & Swin-BIDDS & \textbf{0.9984} & \textbf{0.9698} & \textbf{0.8803} & \textbf{0.9666} & \textbf{0.9906} \\
\hline
MSMT 17 & NLCRIM & 0.9909 & 0.5979 & 0.3114 & 0.5428 & 0.8604 \\
 & LCRIM & 0.9909 & 0.6306 & 0.3609 & 0.5256 & 0.8510 \\
  & BIDDS & 0.9955 & 0.8426 & 0.6466 & 0.7640 & 0.9406 \\
 & Swin-BIDDS & \textbf{0.9993} & \textbf{0.9750} & \textbf{0.9082} & \textbf{0.9445} & \textbf{0.9897} \\
\hline
Outdoor Gait & NLCRIM & 0.9840 & 0.3671 & 0.1513 & 0.8435 & 0.9889 \\
 & LCRIM & 0.9816 & 0.4021 & 0.1550 & 0.8399 & 0.9835 \\
  & BIDDS & 0.9964 & 0.8295 & 0.4672 & 0.9465 & 0.9951 \\
 & Swin-BIDDS & \textbf{0.9992} & \textbf{0.9657} & \textbf{0.7003} & \textbf{0.9802} & \textbf{0.9978} \\
\hline
DeepChange & NLCRIM & 0.8709 & 0.0720 & 0.0176 & 0.1539 & 0.4786 \\
 & LCRIM & 0.8551 & 0.0902 & 0.0276 & 0.1337 & 0.4444 \\
  & BIDDS & 0.8869 & 0.2021 & 0.0897 & 0.2276 & 0.5504 \\
 & Swin-BIDDS & \textbf{0.9861} & \textbf{0.4227} & \textbf{0.1527} & \textbf{0.5028} & \textbf{0.8836} \\
\hline
\end{tabular}
\end{center}
\label{tab:benchmark_results}
\end{table*}

\begin{table}
\caption{BTS Dataset and Partitions}
\begin{center}
\begin{tabular}{|l|c|c|}
\hline
Dataset Partition & IDs & Media Files \\
\hline\hline
\multicolumn{3}{|c|}{Test Sets} \\
\hline
Gallery & 858 & 164,638 \\
Probe & 367 & 9,215 \\
\hline\hline
\multicolumn{3}{|c|}{Probe Subsets} \\
\hline
Face Included & 367 & 5,749 \\
Face Restricted & 367 & 1,893 \\
Long-range Body & 362 & 2,832 \\
Unmanned aerial vehicle & 139 & 834 \\
\hline
\end{tabular}
\end{center}
\label{BTS_summary}
\vskip -.35 cm
\end{table}

\begin{table*}
\caption{BRIAR Test Set performance.}
\begin{center}
\begin{tabular}{|l|l|c|c|c|c|c|}
\hline
Probe Set & Model & AUC & TAR@FAR $10^{-3}$ & TAR@FAR $10^{-4}$ & Rank 1 & Rank 20 \\
\hline\hline
All Probes & NLCRIM & 0.9403 & 0.0698 & 0.0167 & 0.1467 & 0.5825 \\
& LCRIM & 0.9275 & 0.0740 & 0.0188 & 0.1615 & 0.5731 \\
& BIDDS & 0.9745 & 0.2926 & 0.1207 & 0.3342 & 0.7816 \\
& Swin-BIDDS & \textbf{0.9802} & \textbf{0.3575} & \textbf{0.1523} & \textbf{0.3909} & \textbf{0.8228} \\
\hline
Face Included & NLCRIM & 0.9394 & 0.0658 & 0.0183 & 0.1372 & 0.569 \\
& LCRIM & 0.9258 & 0.0699 & 0.0184 & 0.1571 & 0.5655 \\
& BIDDS & 0.9736 & 0.2767 & 0.1089 & 0.3248 & 0.7779 \\
& Swin-BIDDS & \textbf{0.9797} & \textbf{0.3451} & \textbf{0.1388} & \textbf{0.3799} & \textbf{0.8201} \\
\hline
Face Restricted & NLCRIM & 0.9326 & 0.0666 & 0.0116 & 0.1337 & 0.5473 \\
& LCRIM & 0.9173 & 0.0666 & 0.0127 & 0.1432 & 0.5346 \\
& BIDDS & 0.969 & 0.2573 & 0.1051 & 0.2948 & 0.7517 \\
& Swin-BIDDS & \textbf{0.9745} & \textbf{0.3133} & \textbf{0.1352} & \textbf{0.3391} & \textbf{0.7855} \\
\hline
Long-range Body & NLCRIM & 0.9323 & 0.0639 & 0.018 & 0.1063 & 0.5035 \\
& LCRIM & 0.9123 & 0.0749 & 0.018 & 0.119 & 0.4905 \\
& BIDDS & 0.9625 & 0.2331 & 0.0964 & 0.2549 & 0.7023 \\
& Swin-BIDDS & \textbf{0.9685} & \textbf{0.2669} & \textbf{0.1095} & \textbf{0.2832} & \textbf{0.7376} \\
\hline
UAV & NLCRIM & 0.9295 & 0.0779 & 0.0108 & 0.1367 & 0.5468 \\
& LCRIM & 0.9253 & 0.1031 & 0.0228 & 0.1823 & 0.5408 \\
& BIDDS & \textbf{0.9779} & 0.3549 & 0.1835 & 0.3573 & 0.7926 \\
& Swin-BIDDS & 0.9777 & \textbf{0.3765} & \textbf{0.2338} & \textbf{0.4053} & \textbf{0.801} \\
\hline
\end{tabular}
\end{center}
\label{tab:bts_results}
\end{table*}

\subsubsection{Additional Pre-training}\label{LCRIM_pretraining} 
Only LCRIM had additional
pre-training. This comprised a specialized linguistic pre-training phase using the HumanID \cite{o2005video} and MEVA \cite{Corona_2021_WACV} datasets. The HumanID dataset provides diverse viewing scenarios of 297 identities, including approach sequences, walking perpendicular to the camera, and elevated viewpoints. Each identity was annotated by 20 human observers using 30 standardized body descriptors, with the final descriptors averaged across annotators (cf., \cite{myers2023recognizing}). The MEVA dataset, comprising over 9,300 hours of video across varied activities and scenarios, contributed an additional 158 identities. Images from these datasets were used to train LCRIM's initial ability to map between visual features and linguistic body descriptions.
The model was subsequently tuned for mapping
image to identity using the datasets listed 
in Table \ref{training_datasets}.

\subsubsection{Additional Fine-tuning} 
Subsequent to the large-scale training using the datasets in Table \ref{training_datasets}, both the BIDDS and Swin-BIDDS were fine-tuned using the BRIAR training data
(BRS1--5), which contained 697,348 images of 995 unique identities. 
Note: this training data was included in the large scale training and
repeated in the fine-tune stage. During this fine-tuning, BIDDS processes images at an increased size of $384\times384$, allowing for more detailed feature extraction. The Swin-BIDDS model used the $384\times384$ images for both the
large-scale training and the fine-tuning.

In summary, the strategy across models combines specialized linguistic pre-training, extensive foundation-model training, and targeted fine-tuning to fully  exploit the capabilities of each architectural approach. The processing of video-derived images and standardization of input sizes ensures consistent training conditions across the  models.

\section{EXPERIMENTS}\label{expt_section}

\subsection{Benchmark Dataset Tests} 
\subsubsection{Methods}
The models were evaluated first with 
the test data from four benchmark Re-ID datasets 
(MSMT17 \cite{wei2018person}, MARS \cite{zheng2016mars},
Outdoor Gait \cite{song2017learning}, and
DeepChange \cite{xu2023deepchange}) (See Table \ref{training_datasets}). 
DeepChange is a clothes change database; MSMT17, MARS,
and Outdoor Gait are not.
For MSMT17, MARS, and Outdoor Gait the test data were split into
a gallery (half of the items) and a probe set (remaining items).
Because the data are derived from video, the split was made to 
assure minimally similar gallery and probe items. Images were processed
by the models, and identity templates were formed for gallery items by averaging  embeddings of the 
images for each identity.
Identification was measured by comparing probe image embeddings to the gallery templates.  

For DeepChange, all identities had multiple clothing sets. The original DeepChange dataset, however, used similar clothing for each identity across both the probe and gallery sets. Thus, to ensure that clothing did not become a dominant cue, we restructured the partitioning of the probe and gallery sets. Specifically, we designated a single clothing set for all probe instances and then ensured that each identity’s gallery templates had different clothing.

\subsubsection{Results}
\label{subsec:results}

Table~\ref{tab:benchmark_results} summarizes the verification (TAR@FAR) and identification (Rank) performance for the benchmark datasets. The Swin-BIDDS model performed
best on all metrics and for all
datasets.  
At a general level, the transformer-based models (BIDDS and Swin-BIDDS) performed
more accurately than
the ResNet-based models (NLCRIM and LCRIM). This was consistent
across all metrics and for all datasets. Although comparisons
with benchmarks in the literature
are not always possible (or 
transparent),
the rank 1 performance of NLCRIM, BIDDS and Swin-BIDDS exceeded the
state-of-the-art (SOTA) for MARS (cf. previous Rank 1 SOTA: MARS (.908) \cite{he2021dense}). The Swin-BIDDS exceeded the
SOTA  for MSMT17 (cf. previous Rank 1 SOTA: MSMT (.917) \cite{chen2023beyond}) and DeepChange (previous Rank 1 SOTA (.48) \cite{xu2023deepchange}).

It is worth noting that our models were trained on multiple datasets in which clothing was not a reliable cue 
to identity.  The 
strong performance of  BIDDs and Swin-BIDDS  on the no-clothes-change datasets  
indicates that the models utilize
body shape cues, and other 
identifying information not linked to clothing
(head structure). The strong performance
of the Swin-BIDDS model on DeepChange (clothes-change) is consistent with this conclusion.

\subsection{Identification in Unconstrained Datasets}
\subsubsection{Methods}
Next, the models were evaluated using the most challenging of the datasets. Specifically, we used the BRIAR Test Set (BTS)
summarized in Table \ref{BTS_summary}.  The first test
was conducted on the entire dataset and subsequent tests
were done on targeted partitions of the data into probe items. These partitions included: a.) face-included items; b.) face-restricted items, c.) long-range items taken
at distance, and d.) items captured from overhead using an UAV.  

To test identification, gallery embedding templates were formed by averaging the embeddings across all still images for each identity. 
Probe embedding templates were derived from video segments,  indicating the specific frames to be used from the video. The embedding for each probe was computed by averaging the frame-level embeddings across this subset of frames. As a result, the videos for a given identity each contributed multiple probe embeddings (one per segmented clip). 

\subsubsection{Results}
Table \ref{tab:bts_results}
shows that Swin-BIDDS performed substantially better than the other models on nearly all metrics. The table also shows the consistency of this advantage across the test partitions.  
As for the benchmark datasets, the ViT-based models (BIDDS and Swin-BIDDS) were clearly superior to the  ResNet models.  Although it is difficult
to compare across partitions,  especially given the different numbers
of items in each set, Rank 1 and Rank 20 performance suggest that 
Swin-BIDDS provides consistently strong identity information for 
probes with and without a visible face, probes at a distance, and
probes taken from overhead (UAV). Moreover, 
despite differences in the overall performance  of the four models,
none collapsed on the partition tests, highlighting the diversity and quantity of the training data in the success of the models.  
 


\subsection{Ablation Experiments: Architecture vs.\ Input Size?}
The Swin-BIDDS model performed best on the benchmark datasets
and  the challenging BRIAR data. It was also consistently
best on all partitions of the BRIAR
data.
In these  experiments, we test factors that might account for
the superior performance of Swin-BIDDS over 
its closest competitor, BIDDS. 
The models differed in two ways.
The backbone architecture changed from a ViT model (BIDDS) to a 
Swin-ViT (Swin-BIDDS) and the input image size for  the core
model training changed from
$224\times224$ sized images (BIDDS) to $384\times384$ sized images (Swin-BIDDS).
Both models were fine-tuned with $384\times384$ images.

Technically, changes in architecture and image
size are independent. However, a simultaneous change
in both is not uncommon, due to the fact that the Swin-ViT
scales far more efficiently than ViT with increasing input image  size. It does this by implementing a hierarchical structure of shifting attention windows, giving Swin-ViT the desirable feature of linear complexity as a function of image size \cite{liu2021swin}.
Thus, a change from a ViT to Swin-ViT is often undertaken with 
the goal of minimizing the computational resources
required for an increase in image size. 

\subsubsection{Methods}

To tease apart whether architecture, image size, or both 
were responsible for the performance boost, we trained  additional models as comparators.
To test directly for the effects of image size independent
of architecture, we compared two Swin-ViT models.
The first was the Swin-BIDDS model  tested in the previous
experiments.  This is a Swin-ViT model that used 
image size $384\times384$ for core training and  fine-tuning. For clarity,
we refer to this as
Swin-BIDDS(384,384).
For the comparison model, we  trained Swin-BIDDS(224,224). This
is a Swin-ViT model that used  
$224\times224$ images for core training and fine-tuning.

To determine the impact of the image size independent of architecture 
we compared a ViT model with a Swin-ViT model, keeping image size constant. 
Specifically, the Swin-BIDDS(224,224) model was compared with a
BIDDS(224,224). This latter is a version of the
BIDDS(224,384) model used in previous experiments, but
with the image size for fine-tuning lowered to 224.

\subsubsection{Results}
\begin{figure*}
      \centering
      \includegraphics[width=8cm]{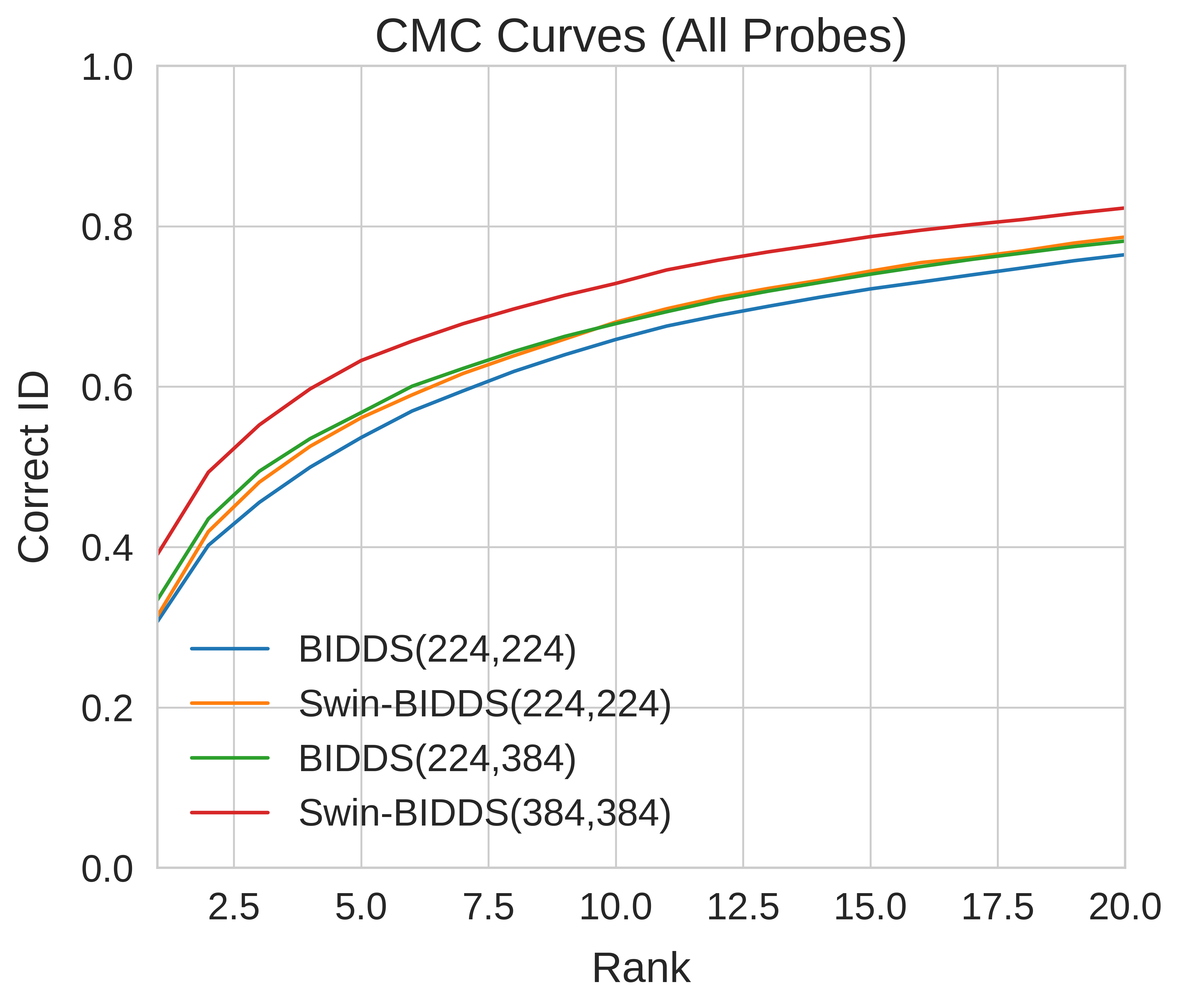}
      \includegraphics[width=8cm]{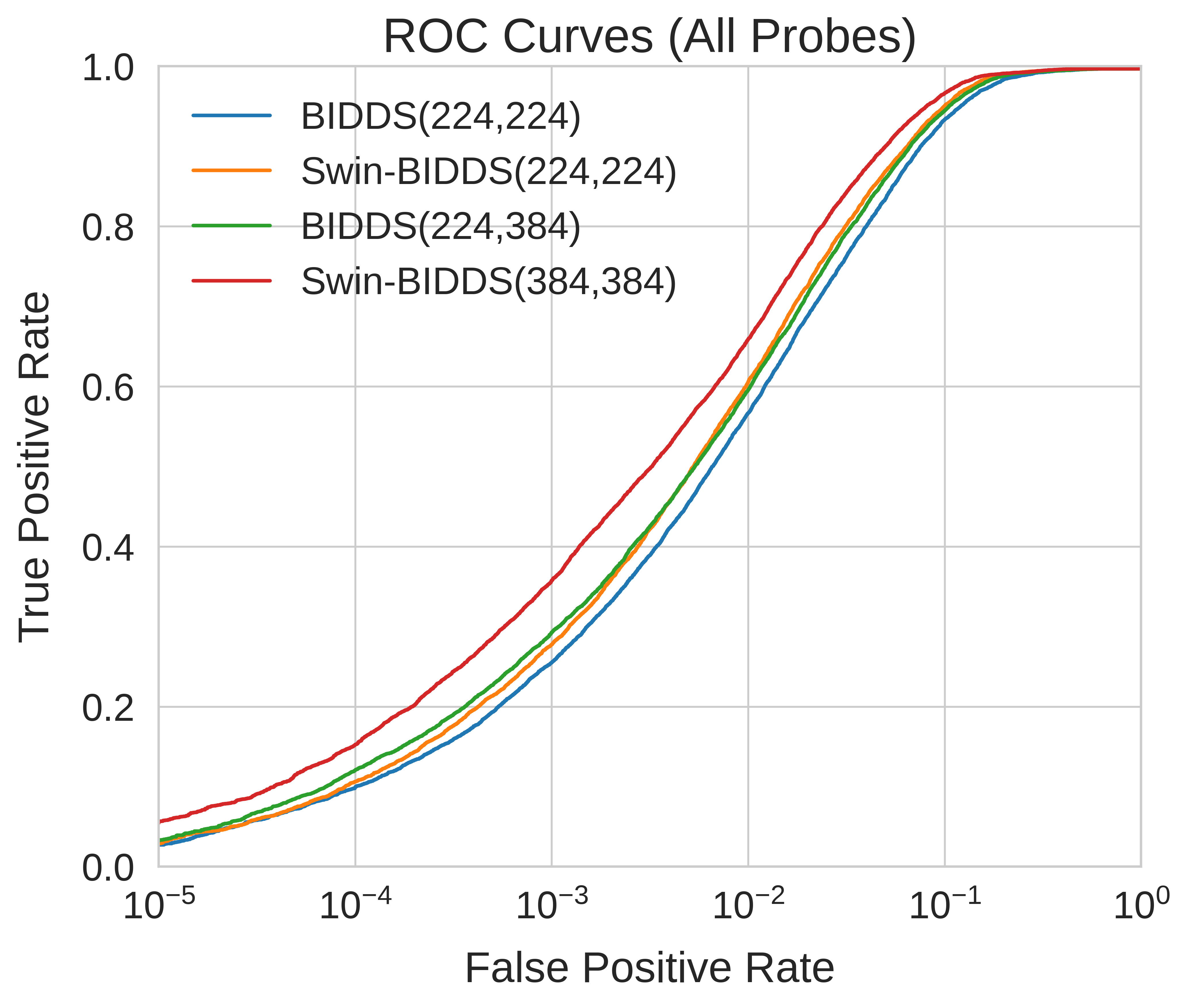}
      \caption{Ablation: CMC and ROC curves for the architecture and image size comparisons show that image size is the critical factor in the superior performance of the Swin-BIDDS model over the BIDDS model.}
      \label{CMC_ROC_BRIAR_all}

   \end{figure*}

Plots showing the CMC and ROC of these models appear in Figure \ref{CMC_ROC_BRIAR_all}. 
Both plots indicate that the performance of the
Swin-BIDDS(384,384) surpasses the other models
 primarily based
on its processing of 
the larger image size. 
There is also a smaller contribution of the Swin-ViT 
architecture, which is  
seen more clearly in Table \ref{ablation_table}. 
Metrics for each model appear in the top rows. 
At the bottom of the table,
the first row shows the effects of changing architecture from the
ViT to Swin-ViT. This is computed as the difference
between the two architecture comparator networks.
All 4 metrics increase with that change.  The final
row of Table \ref{ablation_table} shows 
the effects of changing image size from the smaller 
(224, 224) size to larger (384, 384) size.  
This is computed as the difference
between the two image size comparator networks.
Again, all 4 metrics increase, but
by a substantially larger margin.

\begin{figure*}[htb]
    \centering
    \includegraphics[width=.75\linewidth]{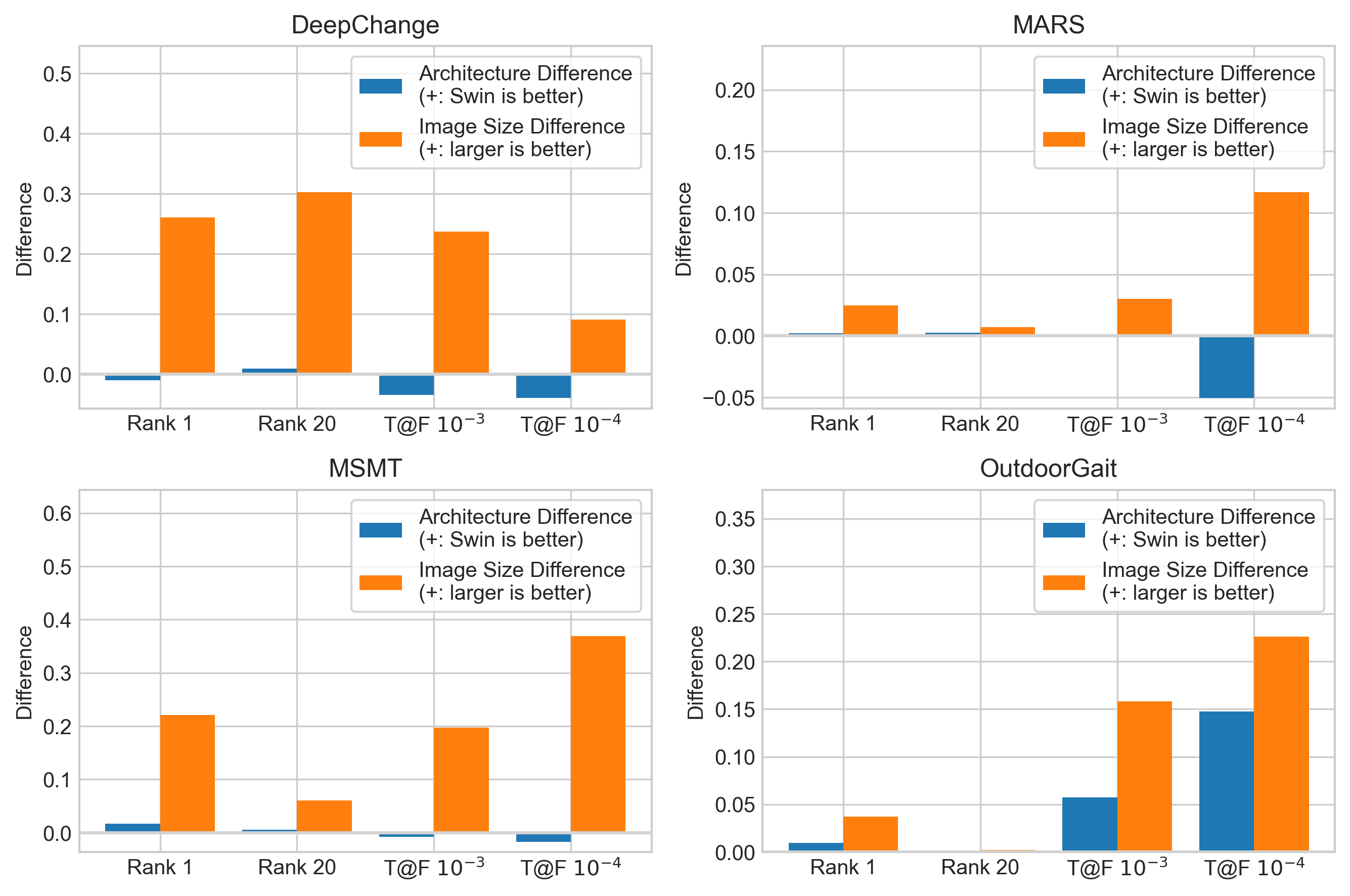}
    \caption{ Difference in performance by model architecture (ViT, Swin-ViT) and input image size (\mbox{224 px$^2$}, \mbox{384 px$^2$}). Comparisons shown for each of four datasets (DeepChange, MARS, MSMT, and Outdoor Gait) on four identification metrics (retrieval at ranks 1 and 20, true accept rate at false accept rates $10^{-3}$ and $10^{-4}$). Architecture Difference (blue) is defined as the difference between Swin-BIDDS(224,224) and BIDDS(224,224). Image Size Difference (orange) is defined as the difference between Swin-BIDDS(384,384) and Swin-BIDDS(224,224). }
    \label{fig:arch_vs_size}
\end{figure*}

As a final check on the consistency of image size  as 
the critical factor in the superior performance of  
Swin-BIDDS over BIDDS, we conducted the same comparisons
on the benchmark datasets.  The results appear in Figure 
\ref{fig:arch_vs_size} and show again that image size is
the driving factor in the superior performance of Swin-BIDDS over
BIDDS. In these benchmark datasets, changing the backbone
architecture had  variable effects for different datasets.

\begin{table*}[htb]
\caption{Ablation Results on BRIAR Test Set, All Probes.}
\begin{center}
\begin{tabular}{|l|c|c|c|c|c|c|}
\hline
Architecture & Core & Fine-Tune & TAR@FAR $10^{-3}$ & TAR@FAR $10^{-4}$ & Rank 1 & Rank 20 \\
\hline\hline
BIDDS & 224 & 224 & 0.2549 & 0.0990 & 0.3072 & 0.7645 \\
Swin-BIDDS & 224 & 224 & 0.2776 & 0.1063 & 0.3141 & 0.7865 \\
BIDDS & 224 & 384 & 0.2926 & 0.1207 & 0.3342 & 0.7816 \\
Swin-BIDDS & 384 & 384 & 0.3575 & 0.1523 & 0.3909 & 0.8228 \\
\hline\hline
Architecture: BIDDS to Swin-BIDDS & & & +0.0227 & +0.0073 & +0.0069 & +0.0220 \\
Image size: 224 to 384 & & & \textbf{+0.0799} & \textbf{+0.0460} & \textbf{+0.0768} & \textbf{+0.0363} \\
\hline
\end{tabular}
\end{center}
\label{ablation_table}
\end{table*}



\section{CONCLUSIONS AND FUTURE WORKS}
We implement four long-term body identification
models based on ResNet (LCRIM, NLCRIM) and 
ViT (BIDDS, Swin-BIDDS) architectures.
The models are tested on their ability to identify bodies
in benchmark re-identification  datasets and in a highly challenging unconstrained dataset that includes people viewed at a distance, from elevated vantage points, and with clothing variability. An important aspect of our approach
is that we train the models on a large-scale, diverse dataset of nearly two million images of nearly 5,000 identities.
We showed that vision transformer architectures consistently outperformed ResNet architectures. Swin-BIDDS was the most accurate model across metrics and models. Its primary advantage over the BIDDS model was its use of a larger input image size. The results underscore the importance of leveraging large image size and hierarchical self-attention in capturing subtle body shape differences. Overall, the Swin-BIDDS model demonstrates strong generalization across benchmark datasets and the BRIAR set, making it a highly robust approach for long-term, real-world body identification.

Promising avenues for future work include the use
of larger and more diverse training sets, and new learning paradigms that leverage unlabeled data in complex viewing conditions. 
Moreover, unsupervised or semi-supervised approaches that exploit partially labeled videos could reduce reliance on fully annotated datasets, potentially handling rare body shapes and challenging occlusions more robustly. Integrating such techniques directly into the current architectures or using them in conjunction with shape-based encoders may open the door to even more accurate and resilient body-identification systems under real-world constraints.

\section{ACKNOWLEDGMENTS}

This research is based upon work supported in part by
the Office of the Director of National Intelligence (ODNI),
Intelligence Advanced Research Projects Activity (IARPA),
via [2022-21102100005]. The views and conclusions contained herein are those of the authors and should not be
interpreted as necessarily representing the official policies,
either expressed or implied, of ODNI, IARPA, or the U.S.
Government. The US. Government is authorized to reproduce
and distribute reprints for governmental purposes notwithstanding any copyright annotation therein.


\section*{ETHICAL IMPACT STATEMENT}
We have read the guidelines for the Ethical Impact Statement.
The development of  body identification models does not 
 involve direct contact with human subjects, and therefore
 does not require approval by an Institutional Review Board. 
 Instead,  images/videos 
of human subjects are incorporated as training 
and test data for body identification models. We used only datasets 
(videos and images of people) that have been pre-screened and approved for ethical data collection standards by a United States government
funding agency, XXXX. The standards applied for dataset approval
require consent from the subjects who are depicted in the images/videos for use in research. Specifically, the standards are set in accordance with Health Services Research and applicable privacy policies, statutes, and federal regulations.
Images/videos of subjects who
appear in publications require additional consent.
We followed these guidelines carefully. Images displayed in
the paper have been properly consented  and are displayed
according to the published  instructions for use of the dataset.

The development and study of biometric identification algorithms entails risk to individuals and societies. It is clear that these systems can have negative impacts if they are
misused. They can potentially threaten
individual  privacy and can impinge on  freedom of movement and expression in a society. 
The goal of our work is to better understand 
how these systems work. The results of this work can have both positive and negative societal 
impacts. On the positive side, knowing the types of representations created by body identification networks can help to minimize person identification errors. It can also help to set reasonable performance expectations thereby limiting the scope of use. On the negative side,  the knowledge gained can potentially be used to manipulate a 
system in unethical ways and to create synthetic images that can be misused or misinterpreted.

These risks are mitigated by the potential for positive societal impact. Body identification algorithms can  be used to locate missing people (including children). They can also be used in law enforcement to identify individuals implicated in crimes. Legitimate and societally-approved use can protect the general public from harm.  Of note, body identification systems can be used in combination with face identification systems to improve identification accuracy, thereby minimizing erroneous identifications.

\addtolength{\textheight}{-3cm}   



{\small
\bibliographystyle{ieee}
\bibliography{references}
}

\end{document}